\documentclass{article}
\usepackage{spconf,amsmath,graphicx}
\usepackage[euler]{textgreek}
\usepackage{ tipa }
\usepackage{ipa}
\usepackage{amsmath}

\usepackage{savesym}
\savesymbol{eth}

\usepackage{diagbox}
\usepackage{bm}

\usepackage{amssymb}
\restoresymbol{TXF}{eth}

\usepackage{amsfonts}

\usepackage{multirow}
\usepackage{float}
\usepackage{hyperref}

\title{Multilingual and crosslingual speech recognition using phonological-vector based phone embeddings}
\name{Chengrui Zhu, Keyu An, Huahuan Zheng, Zhijian Ou$^{\dagger}$\thanks{$\dagger$ Corresponding author. This work is supported by NSFC 61976122. Code is available at https://github.com/thu-spmi/CAT}}
\address{Speech Processing and Machine Intelligence (SPMI) Lab, Tsinghua University, China\\
Beijing National Research Center for Information Science and Technology, China }
\begin{document}
\ninept
\maketitle
\begin{abstract}

The use of phonological features (PFs) potentially allows language-specific phones to remain linked in training, which is highly desirable for information sharing for multilingual and crosslingual speech recognition methods for low-resourced languages.
A drawback suffered by previous methods in using phonological features is that the acoustic-to-PF extraction in a bottom-up way is itself difficult.
In this paper, we propose to join phonology driven phone embedding (top-down) and deep neural network (DNN) based acoustic feature extraction (bottom-up) to calculate phone probabilities.
The new method is called JoinAP (Joining of Acoustics and Phonology). Remarkably, no inversion from acoustics to phonological features is required for speech recognition.
For each phone in the IPA (International Phonetic Alphabet) table, we encode its phonological features to a phonological-vector, and then apply linear or nonlinear transformation of the phonological-vector to obtain the phone embedding.
A series of multilingual and crosslingual (both zero-shot and few-shot) speech recognition experiments are conducted on the CommonVoice dataset (German, French, Spanish and Italian) and the AISHLL-1 dataset (Mandarin), and demonstrate the superiority of JoinAP with nonlinear phone embeddings over both JoinAP with linear phone embeddings and the traditional method with flat phone embeddings.

\end{abstract}
\section{Introduction}
\label{sec:intro}

In recent years, deep neural network (DNN) based automatic speech recognition (ASR) systems have been improved dramatically, which are, however, data-hungry.
A well-trained DNN based ASR system for a single language usually requires hundreds to thousands of hours of transcribed speech data.
Remarkably, there are more than 7100 languages in the world \cite{lewis}, and most of them are low-resourced languages, for which only limited transcribed speech data are available \cite{GlobalPhone}.

To advance ASR for low-resourced languages, multilingual and crosslingual speech recognition methods have long been developed, mainly for acoustic modeling \cite{Schultz98,Heigold,10.1109/TASLP.2015.2422573,tandem_is2010,Hermann_2018,idiap_arxiv17,Idiap_IS18} (More discussions in Section \ref{sec:related}). 
Some end-to-end ASR models \cite{asru17,g_ic18,g_ic19} fold the acoustic model (AM), pronunciation lexicon and language model (LM) into a single neural network, making the models being even more data-hungry and not suitable for low-resourced multilingual speech recognition, which is the main problem we hope to solve in this work\footnote{We suppose that text corpus and pronunciation lexicons or grapheme-to-phoneme (G2P) transducers \cite{Mark_slsp20,novak-etal-2012-wfst} are available.}.

\begin{figure}[t]
\centering
\centerline{\includegraphics[width=6.5cm]{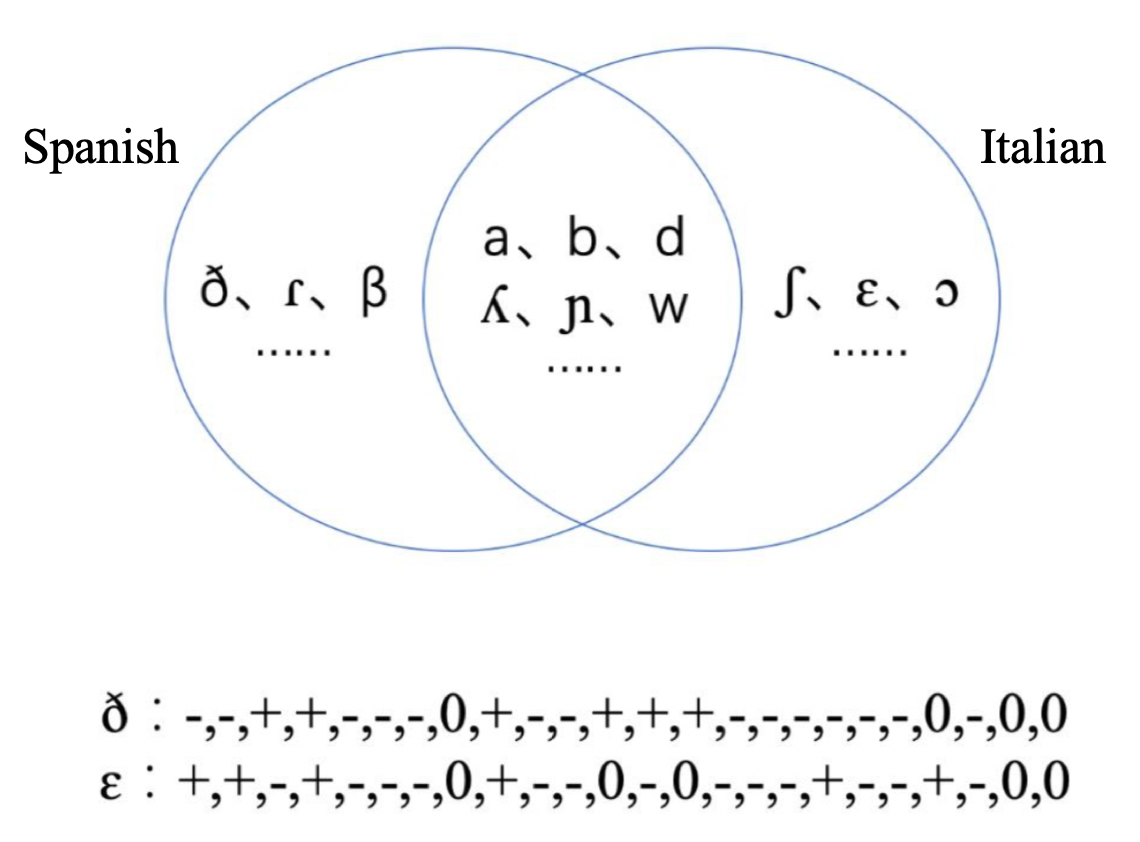}}
\vspace{-3mm}
\caption{Illustration of the connection between Spanish and Italian. As indicated by the intersection of the two circles, there are some common IPA phones used in both languages, which can be trained with more data from both languages. 
Notably, there also exist language-specific phones for each language. \eth~and~\textepsilon~only appears in Spanish and Italian respectively, and thus are not linked in the surface forms.
The bottom show the phonological features for~\eth~and~\textepsilon, which share many common components. Phonological feature components are ordered as listed in Table \ref{tab:PFfeature}.}
\label{fig:connect}
\vspace{-4mm}
\end{figure}

Intuitively, the key to successful multilingual and crosslingual recognition is to promote the information sharing in multilingual training and maximize the knowledge transferring from the well trained multilingual model to the model for recognizing the utterances in the new language.
To this end,  a common practice is that similar sounds across languages are combined into one multilingual phone set. 
International Phonetic Alphabet (IPA), which classifies sounds based on phonetic knowledge, has been used to create a universal phone set \cite{Schultz98,idiap_arxiv17,2020That}. 
Often phones are seen as being the ``atoms'' of speech. But it is now widely accepted in phonology that phones are decomposable into smaller, more fundamental units, sharable across all languages, called phonological distinctive features \cite{nla.cat-vn542882,King_csl00}.
Namely, phones can be represented by a set of phonological features, such as voicing, high, low (representing tongue position during vowels), round (for lip rounding), continuant (to distinguish sounds such as vowels and fricatives from stops) and so on, as shown in Table \ref{tab:PFfeature}.
As shown in Fig. \ref{fig:connect},  the use of phonological features potentially allow language-specific phones to remain linked — to ``share statistical strength'' in training. This is highly desirable, especially for zero-shot crosslingual speech recognition.

\begin{figure}[t]
\centering
\centerline{\includegraphics[width=8cm]{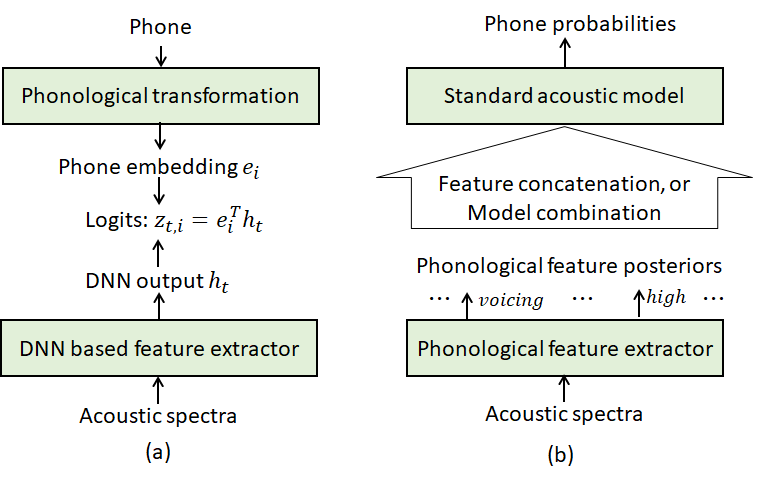}}
\vspace{-3mm}
\caption{(a) Phonology driven phone embedding (top-down) and DNN based acoustic feature extraction (bottom-up) are joined to calculate the logits, which define the phone probabilities.
(b) Traditional methods in using phonological features are purely bottom-up.}
\label{fig:methods}
\vspace{-3mm}
\end{figure}

Phonological features (PFs) have been applied in multilingual and crosslingual speech recognition \cite{cmu_EU03,Idiap_IS18}.
Previous studies generally take a bottom-up approach and train a phonological feature extractor, often implemented by neural networks.
Each training sample consists of a speech frame as the input and the canonical phonological feature components derived from the labeled phone as the target output values.
Multiple neural networks are trained, depending on the partition of the phonological features.
In the feature concatenation approach, the log posteriors of every phonological class are concatenated together and fed into the high-level acoustic model, which further predicts the phone probabilities \cite{Idiap_IS18}. 
Alternatively, in the model combination approach, the PF probabilities and the phone probabilities from a standard acoustic model are combined to calculate the acoustic score \cite{cmu_EU03}.

A drawback suffered by previous methods in using phonological features is that the acoustic-to-PF extraction in a bottom-up way is itself difficult, let alone the training of the phonological feature extractor needs segmented and labeled speech at the phone level.
Moreover, previous methods do not provide a principled model to calculate the phone probabilities for unseen phones from the new language towards zero-shot crosslingual recognition.
The parameters connecting to the unseen phones in the output layer of the DNN model are initialized either randomly \cite{Schultz98} or in an ad-hoc way (taking a weighted average of the parameters of all the seen phones \cite{Idiap_IS18}).

In this paper, we propose a new approach to using phonological features for multilingual and crosslingual speech recognition.
As illustrated in Fig. \ref{fig:methods}(a), our approach consists of phonology driven phone embedding (top-down) and DNN based acoustic feature extraction (bottom-up), which are joined to calculate the logits to define the phone probabilities. This is different from the pure bottom-up manner of the traditional methods in using phonological features, as sketched in Fig. \ref{fig:methods}(b).

Specifically, by using binary encoding of phonological features, we first obtain an encoding vector for each phone in the IPA table, which is referred to as the \emph{phonological-vector}. Then, we apply linear or nonlinear transformation of the phonological-vector to obtain the phone embedding vector for each phone. This step is referred to as the phonological transformation (top-down). Next, we conduct bottom-up calculation on an acoustic DNN, viewed as a cascade of acoustic feature extractors.
Finally, the extracted acoustic features and the phone embeddings are joined to calculate the phone (posterior) probabilities, which can be further used to calculate the CTC loss \cite{Graves_CTC} or the CTC-CRF loss \cite{CTC-CRF-icssp2019,an2020cat}. This completes the definition of a multilingual acoustic model, which involves the Joining of Acoustics and Phonology and is thus called the JoinAP method.
Remarkably, no inversion from acoustics to phonological features is required for speech recognition.
Details about applying the JoinAP model to multilingual and crosslingual speech recognition are given in Section \ref{sec:method}.

To evaluate the JoinAP model, a series of multilingual and crosslingual (both zero-shot and few-shot) speech recognition experiments are conducted on the CommonVoice dataset (involving German, French, Spanish and Italian) \cite{ardila2020common} and the AISHLL-1 dataset (Mandarin) \cite{bu2017aishell1}. The main findings are as follows. 
\begin{itemize}
    \item With JoinAP, we can develop a single acoustic model for multilingual speech recognition, which performs better than the traditional multilingual model (namely using flat phone embeddings\footnote{In the traditional multilingual model such as based on CTC or CTC-CRF, the weights of the final linear layer before softmax can be regarded as flat, unstructured phone embeddings. In contrast, JoinAP uses phonological-vector based, structured phone embeddings.}).
    \item In zero-shot crosslingual recognition, JoinAP with nonlinear phone embeddings outperforms both JoinAP with linear phone embeddings and the traditional model with flat phone embeddings significantly and consistently.
    \item In few-shot crosslingual recognition, using JoinAP with nonlinear phone embeddings still yields much better results than using flat phone embeddings; however, the superiority of nonlinear over linear for phone embeddings seems to be weakened, as there are more training data from the target languages.
\end{itemize}

\section{Related Work}
\label{sec:related}

In \emph{multilingual speech recognition}, training data for a number of languages, often referred to as seen languages, are merged to train a multilingual AM. 
Multilingual training is found to outperform monolingual training, which trains the monolingual AMs separately for each seen language \cite{Heigold}. Such advantage is presumably due to the information sharing between seen languages in multilingual training.
A common approach is to share the lower layers of the DNNs between languages, while the output layers are language specific \cite{Heigold,10.1109/TASLP.2015.2422573}.
Another widely used approach is to extract the bottleneck features from the bottleneck layer of a multilingual DNN model, which are then used as input features to train the AM for the target language \cite{tandem_is2010,Hermann_2018}.

\emph{Crosslingual speech recognition} refers to recognizing utterances in a new language, which is unseen in training the multilingual AM.
In the zero-shot setting, the multilingual AM is trained and directly used without any transcribed speech from the new, target language \cite{Schultz98,2020That,2020How,LXJ_AAAI2020}.
Alternatively, in the few-shot setting, the multilingual AM can be further finetuned or adapted on limited transcribed speech from the new language \cite{idiap_arxiv17,Idiap_IS18,li2020universal}.
Hopefully, knowledge can be transferred, by adapting a well-trained multilingual model to the new language, presumably because the multilingual model should learn some universal phonetic representations and the new language is similar to seen languages, more or less.

Earlier studies in multilingual and crosslingual recognition use context-dependent phone units, which leads to an explosion of units and also needs special care to handle context-dependent modeling across languages \cite{eu01-triphone,Idiap_Ic19_LFMMI}.
There are recent attempts to use end-to-end ASR models such as CTC with monophones \cite{idiap_arxiv17,LXJ_AAAI2020} or end-to-end LF-MMI with biphones \cite{E2E-LF-MMI,Idiap_Ic19_LFMMI} for multilingual and crosslingual recognition.
Remarkably, the end-to-end CTC-CRF model, which is defined by a CRF (conditional random field) with CTC topology, has been shown to perform significantly better than CTC \cite{CTC-CRF-icssp2019,an2020cat}. Moreover, mono-phone CTC-CRF performs comparably to bi-phone end-to-end LF-MMI \cite{E2E-LF-MMI} and avoids context-dependent modeling with a simpler pipeline, which is particularly attractive for multilingual and crosslingual speech recognition.

When the phonological transformation is linear, our JoinAP model reduces to the model introduced in \cite{LXJ_AAAI2020}. But in \cite{LXJ_AAAI2020}, only zero-shot crosslingual phone recognition is conducted and the model is developed still in a bottom-up way without the idea of joining acoustics (bottom-up) and phonology (top-down).

\section{Method}
\label{sec:method}

This section first explains the definition and construction of phonological vectors. Then we describe the JoinAP method with linear and nonlinear phone embeddings. Finally, we introduce the CTC-CRF based ASR framework to use JoinAP.

\subsection{Phonological-vector}
\label{ssec:phononlogical}

Phonological (distinctive) features have been proposed as the basis of spoken language universals, in the sense that while the phones of a language vary, the set of phonological features does not and is the same for all languages.
That is, phones can be constructed from a set of phonological features.
As shown in Fig. \ref{fig:connect},  the use of phonological features potentially allow language-specific phones to remain linked, which could benefit the information sharing in multilingual training.

There are different phonological feature sets and phonological systems, among which one of the most popular systems is proposed by Chomsky and Halle in 1968~\cite{nla.cat-vn542882}.
Phonological features are categorized into four classes: major class features, manner of articulation features, source features, cavity features. Each feature is marked as `+', `-' or `0'. `+' indicates the presence of that feature, `-' indicates the absence, and `0' means certain phone does not show such feature; for example, it is meaningless for a vowel to possess consonant features, so it will be marked as `0'. In our experiment, we employ PanPhon \cite{mortensen-etal-2016-panphon} to obtain the phonological features for IPA symbols. PanPhon uses a total of 24 phonological features. Table \ref{tab:PFfeature} gives examples of the feature specifications of some IPA phones, where all 24 features are listed.

\begin{table}[t]
\centering
\caption{Phonological features of some IPA phones}
\label{tab:PFfeature}
\begin{tabular}{l|lllllll}
\hline
Phonological feature           & d  & \textepsilon & \eth & \textschwa & i &  \textdctzlig & k\textsuperscript{j}\\
\hline
syllabic          & -  & + & - & + & + &- &-\\
sonorant          & -  & + & - & + & + & -&-\\
consonantal       & +  & - & + & - & - & +&+\\
continuant        & -  & + & + & + & + & -&-\\
delayed release   & -  & - & - & - & - & +&-\\
lateral           & -  & - & - & - & - & -&-\\
nasal             & -  & - & - & - & - & -&-\\
strident          & 0  & 0 & 0 & 0 & 0 & 0&0\\
voice             & +  & + & + & + & + & +&-\\
spread glottis    & -  & - & - & - & - & -&-\\
constricted glottis & -  & - & - & - & - & -&-\\
anterior          & +  & 0 & + & 0 & 0 & -&-\\
coronal           & +  & - & + & - & - & +&-\\
distributed labial & -  & 0 & + & 0 & 0 & +&0\\
labial            & -  & - & - & - & - & -&-\\
high              & -  & - & - & - & + & +&+\\
low               & -  & - & - & - & - & -&-\\
back              & -  & - & - & + & - & -&-\\
round             & -  & - & - & - & - & -&-\\
velaric        & -  & - & - & - & - & -&-\\
tense             & 0  & - & 0 & - & + & 0&0\\
long             & -  & - & - & - & - & -&-\\
hitone             & 0  & 0 & 0 & 0 & 0 &0 &0\\
hireg             & 0  & 0 & 0 & 0 & 0 & 0&0\\
\hline
\end{tabular}
\vspace{-3mm}
\end{table}


Now each phone is described by 24 phonological features, and each feature can take `+', `-' or `0'. Further, we encode each phonological feature by a 2-bit binary vector. 
Taking the feature ``round'' as an example, the first bit indicates whether it is ``round+'' and the second bit indicates ``round-''. Therefore, if the ``round'' feature takes `+', the 2-bit vector will be ``10''; if the ``round'' feature is `-', the 2-bit vector will be ``01''; if the ``round'' feature takes `0', the 2-bit vector will be ``00''. 
In this way, we can represent the phonological features by a 48-bit vector. Additionally, acoustic training (e.g., based on CTC-CRF) introduces 3 special extra tokens (\verb|<blk>|, \verb|<spn>| and \verb|<nsn>|), so we further add another 3 bits to encode the three special tokens in one-hot.
In summary, we obtain a 51-bit encoding vector for each phone in the IPA table, which is referred to as the \emph{phonological-vector}.

\subsection{Phone embedding}

Based on the phonological-vector representation of phones, we propose to join phonology driven phone embedding (top-down) and DNN based acoustic feature extraction (bottom-up) to calculate the logits, which further define the phone probabilities for ASR.
This method, called JoinAP (Joining of Acoustics and Phonology), is different from the traditional bottom-up way to use phonological features for ASR (e.g., by building the phonological feature extactor) (See Fig. \ref{fig:methods} for illustration).

In the traditional multilingual model such as based on CTC or CTC-CRF, one may view the acoustic DNN as a cascade of bottom-up feature extractors.
At frame $t$, the DNN output $h_t \in \mathbb{R}^{H}$ could be viewed as the projection of speech into some abstract space, pertaining to the spoken phones.
Before softmax computation to output phone probabilities, the final linear layer calculates the logits as follows:
\begin{equation}
    \label{eq:linear}
    z_{t,i} = e_i^T h_t
\end{equation}
where $e_i \in \mathbb{R}^{H} $ denotes the weight vector in the final linear layer and could be viewed as a (flat) phone embedding vector for phone $i$.
For simplicity, we omit the bias in describing linear layers throughout the paper.

In JoinAP, we propose to apply linear or nonlinear transformation of the phonological-vector to obtain the phone embedding vector for each phone. This step is referred to as the phonological transformation (top-down) and explained as follows.

\paragraph*{The JoinAP-Linear method.} Given the phonological-vector $p_i \in \mathbb{R}^{51}$ for phone $i$, we apply linear transformation of $p_i$ to define the embedding vector for phone $i$:
\begin{equation}
    \label{eq:embedding}
    e_i=A p_i \in \mathbb{R}^{H}
\end{equation}
where $A \in \mathbb{R}^{H\times51}$ denotes the transformation matrix. The logits for calculating the phone (posterior) probabilities are still defined as in Eq. (\ref{eq:linear}), which can be transparently used in the CTC-CRF based ASR framework (detailed later).


\paragraph*{The JoinAP-Nonlinear method.} The nonlinear method is similar to the linear one, except that we apply nonlinear transformation of $p_i$ to define the embedding vector for phone $i$. In theory, multilayered neural networks could be used for phonological transformation. Here we consider to add one hidden layer as follows:
\begin{equation} \label{eq:nonlinear}
    e_i = A_2 \sigma(A_1 p_i) \in \mathbb{R}^H
\end{equation}
where $A_1, A_2$ denote the matrices of appropriate sizes, and $\sigma(\cdot)$ denote some nonlinear activation function (e.g. sigmoid).
The logits for calculating the phone (posterior) probabilities are still defined as in Eq. (\ref{eq:linear}).



\subsection{CTC-CRF based ASR}
In this section, we briefly explain the CTC-CRF based framework \cite{CTC-CRF-icssp2019, an2020cat} to use phone embeddings for ASR. Consider discriminative training with the objective to maximize the conditional likelihood~\cite{CTC-CRF-icssp2019}:
\begin{equation} 
    \label{eq:crf-obj1}
    \mathcal{L}(\theta) = - \log p_{\theta}(\bm{l}|\bm{x})
\end{equation}
where $\bm{x} \triangleq (x_1, \cdots, x_T)$ is the speech feature sequence, $\bm{l} \triangleq (l_1, \cdots, l_L)$ is the phone-label sequence, and $\theta$ denotes the model parameters. Note that in speech recognition, $\bm{x}$ and $\bm{l}$ are in different lengths and not aligned. To handle this, a hidden state sequence $\bm{\pi} \triangleq (\pi_1,\cdots,\pi_T)$ and a map $\mathcal{B}(\cdot)$ for mapping $\bm{\pi}$ to $\bm{l}$ are introduced. The mapping function $\mathcal{B}$ removes consecutive repetitive labels and blanks in $\bm{\pi}$ to give $\bm{l}$. So the posterior of $\bm{l}$ is defined as:
\begin{equation}
    \label{eq:post-l} 
	p_{\theta}(\bm{l} | \bm{x}) = \sum_{\bm{\pi} \in \mathcal{B}^{-1}(\bm{l})} p_{\theta}(\bm{\pi} | \bm{x})
\end{equation}
And the posterior of $\bm{\pi}$ is further defined by a conditional random field (CRF):
\begin{equation}
    \label{eq:post-pi}
    p_{\theta}(\bm{\pi}|\bm{x}) = 
    \frac{\exp(\phi_{\theta}(\bm{\pi}, \bm{x}))}
    {\sum_{\bm{\pi}'}{\exp(\phi_{\theta}({\bm{\pi}', \bm{x}}))}}
\end{equation}
where $\phi_{\theta}(\bm{\pi}, \bm{x})$ denotes the potential function of the CRF, defined as:
\begin{equation} \label{eq:potential}
    \phi_{\theta}(\bm{\pi}, \bm{x}) = \log p(\bm{l})+ \sum_{t=1}^{T} \log p_{\theta}(\pi_t|\bm{x})
\end{equation}
where $\bm{l} = \mathcal{B}(\bm{\pi})$, and $p(\bm{l})$ is realized by an n-gram LM of labels. If $\log p(\bm{l})$ is omitted in Eq. (\ref{eq:potential}), the potential function becomes self-normalized and CTC-CRF reduces to regular CTC.
$p_{\theta}(\pi_t|\bm{x})$ represents the phone (posterior) probabilities, which are calculated by softmax from the logits $z_{t,i}$ in Eq. (\ref{eq:linear}) as follows:
\begin{displaymath}
p_{\theta}(\pi_t = i|\bm{x}) = \frac{\exp(z_{t,i})}{\sum_j \exp(z_{t,j})}
\end{displaymath}
%

Remarkably, regular CTC suffers from the conditional independence between the states in $\bm{\pi}$. In contrast, by incorporating $\log p(\bm{l})$ into the potential function in CTC-CRF, this drawback is naturally avoided.
It has been shown that CTC-CRF outperforms regular CTC consistently on a wide range of benchmarks, and is on par with other state-of-the-art end-to-end models~\cite{CTC-CRF-icssp2019, an2020cat, CTC-CRF-NAS}. Moreover, CTC-CRF enjoys data-efficiency in training and works well with mono-phones \cite{an2020cat}, which are favorable for low-resourced multilingual and crosslingual speech recognition.

For decoding, we build a weighted finite state transducer (WFST), obtained by composing the CTC topology, pronunciation lexicon and word-level n-gram language model, and use WFST-based decoding. 

\section{Experiment}
\label{sec:exp}

\subsection{Experiment dataset and setup}
Our experiments are conducted on two datasets, CommonVoice \cite{ardila2020common} and AISHELL-1 \cite{bu2017aishell1}. 
In our experiment, we use German, French, Spanish and Italian from CommonVoice to train the multilingual models. We carry out zero-shot and few-shot crosslingual experiments on Polish and Mandarin, where Polish comes from CommonVoice and Mandarin comes from AISHELL-1. Detailed data statistics are shown in Table \ref{tab:datainfo}.

\begin{table}[]
\centering
\caption{Datasets used in our experiments: the source, the number of IPA phone tokens in every language, the size of train, development and test sets in hours.}
\label{tab:datainfo}
\resizebox{0.48\textwidth}{!}{\begin{tabular}{cccccc}

\hline
Language           & Corpora  & \#Phones & Train & Dev & Test\\
\hline
German          & CommonVoice  & 40 & 639.4 & 24.7 & 25.1\\
French          & CommonVoice  & 57 & 465.2 & 21.9 & 23.0\\
Spanish       & CommonVoice  & 30 & 246.4 & 24.9 & 25.6\\
Italian        & CommonVoice  & 33 & 89.3 & 19.7 & 20.8\\
Polish   & CommonVoice  & 46 & 93.2 & 5.2 & 6.1\\
Mandarin           & AISHELL-1  & 96 & 150.9 & 18.1 & 10.0\\
\hline

\end{tabular}}
\end{table}
We employ Phonetisaurus \cite{novak-etal-2012-wfst}, a WFST-based G2P toolkit to generate IPA lexicons for the 6 languages in our experiments.
All the monolingual phones were mapped to IPA symbols and we merged the phones from German, French, Spanish and Italian to create the universal phone
set for multilingual training.

We use the CTC-CRF based ASR Toolkit - CAT \cite{an2020cat}, and will release the code in CAT when this work is published. 
Unless otherwise stated, the acoustic models used in our experiments are all based on CTC-CRF, and word-level N-gram language models are trained on the training transcripts for each language.

In all experiments, 40 dimension filter bank with delta and delta-delta features are extracted as input to the AM, which is 3 blocks of VGG layers followed by a 3-layer BLSTM with 1024 hidden size (namely $H=2048$).
A dropout probability of 50\% is applied to the LSTM to prevent overfitting.
During training, we use Adam as optimizer, and set initial learning rate as 1e-3. When the performance on development set stops improving, learning rate is adjusted to 1/10 of the previous one until it is less than 1e-5.

\subsection{Experiment results}
Our experiments are divided into 2 parts, multilingual and (zero-shot and few-shot) crosslingual. The multilingual models trained on the collection of German, French, Spanish and Italian data are tested on these 4 languages for multilingual experiments and on Polish and Mandarin for crosslingual experiments.

\begin{table*}[]
\centering
\caption{Word error rate (WER) results (\%) for German, French, Spanish and Italian in the multilingual experiments.
Multilingual models are trained with 3 methods: the traditional method using flat phone embeddings (``Flat-Phone''), JoinAP with linear phone embeddings (``JoinAP-Linear''), and JoinAP with nonlinear phone embeddings (``JoinAP-Nonlinear'').
The multilingual models can be directly used without finetuning or with finetuning over the training data of the target languages.
Monolingual models are trained separately for comparisons.}
\label{tab:multi}
\begin{tabular}{cccccccc}
\hline
Language           & Flat-Phone  & Flat-Phone  & Flat-Phone & JoinAP-Linear & JoinAP-Linear & JoinAP-Nonlinear  & JoinAP-Nonlinear\\
                   &  monolingual  & w/o finetuning  & finetuning  & w/o finetuning & finetuning & w/o finetuning & finetuning \\
\hline
German          & 13.09  & 14.36 & 12.42 & 13.72 & 12.45 & 13.97 & 12.64\\
French          & 18.96  & 22.73 & 18.91 & 22.73 & 19.54 & 22.88 & 19.62\\
Spanish       & 15.11  & 13.93 & 13.06 &13.93 & 13.19 & 14.10 & 13.26\\
Italian        & 24.57  & 25.97 & 21.77 & 25.85 & 21.70 & 24.06 & 20.29\\
\hline
Average        & 17.93  & 19.25 & 16.54 & 19.06 & 16.72 & 18.75 & 16.45\\
\hline
\end{tabular}
\vspace{-3mm}
\end{table*}

\begin{table}[]
\centering
\vspace{-1mm}
\caption{About the intersections of the set of phones across languages.
For each phone in a language, we count how many languages it appears and define this count to the language-degree of this phone, which may take from 1 to 4. The cell in column $j$ denotes the number of those phones, whose language-degree is $j=1,2,3,4$.
}
\label{tab:overlapp}
\begin{tabular}{ccccc}
\hline
\diagbox{Language}{Language-degree}     & 4  & 3 & 2 & 1\\
\hline
German          & 18  & 6 & 8 & 8 \\
French          & 18  & 6 & 7 & 26\\
Spanish       & 18  & 4 & 1 & 7 \\
Italian        & 18  & 5 & 4 & 6\\
\hline
\end{tabular}
\vspace{-3mm}
\end{table}

\subsubsection{Multilingual experiment}

Multilingual results are summarized in Table \ref{tab:multi}. 
Multilingual acoustic models are trained with 3 methods: the traditional method (namely using flat phone embeddings), JoinAP with linear phone embeddings, and JoinAP with nonlinear phone embeddings (the hidden layer size is 512). For each method, we test on the target language before and after fine-tuning over the data from the target language.
Monolingual models on German, French, Spanish and Italian are trained separately for comparisons. The main observations are as follows.

\emph{Without finetuning}, the trained multilingual model can be directly used and works as a single model. In this case, on average, both JoinAP-Nonlinear and  JoinAP-Linear perform better than Flat-Phone, and JoinAP-Nonlinear is the strongest. The average relative gain of JoinAP-Nonlinear over Flat-Phone is 3\%. But notably, the detailed improvements of JoinAP-Linear over Flat-Phone and of JoinAP-Nonlinear over JoinAP-Linear are in fact language-dependent.
For Italian, JoinAP-Nonlinear improves the most over JoinAP-Linear (7\%), while for other languages, JoinAP-Linear performs slightly better. JoinAP-Linear performs better or equally well, compared to Flat-Phone.
As observed in previous studies \cite{2020That,Heigold}, the performance differences between different multilingual training methods are affected by several factors, including the phonetic variety in this particular mix of multiple languages, data-scarce/data-rich for the target languages, etc. 

\begin{figure*}[]
\centering
\centerline{\includegraphics[width=18cm]{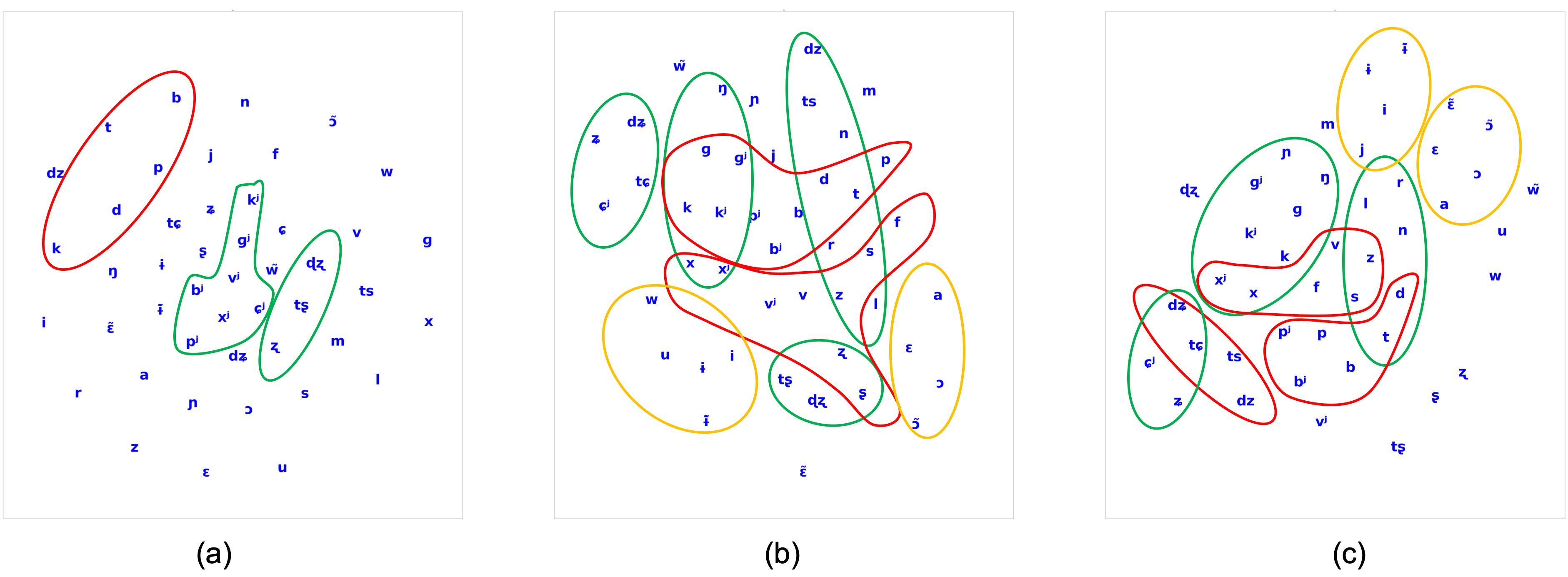}}
\vspace{-3mm}
\caption{Visualization of Polish phone embeddings by t-SNE. (a) Flat phone embeddings, (b) JoinAP-Linear phone embeddings, (c) JoinAP-Nonlinear phone embeddings. They are obtained from the un-finetuned multilingual models in the zero-shot Polish experiment. Red circles indicate the consonants with the same manner of articulation, green circles indicate the consonants with the same place of articulation, and yellow circles indicate similar vowel height.} 
\label{fig:tsne}
\vspace{-3mm}
\end{figure*}

\emph{After finetuing} over the entire data from the target language, we obtain separate models for each target language. 
Similar to previous studies, the three multilingual training methods all significantly outperform monolingual trained models, on average.
JoinAP-Nonlinear reduces the average WER by 8\% against the monolingual models.
The performance differences between the three multilingual methods themselves become smaller, presumably because the target training data are already rich enough to train models. On average, JoinAP-Nonlinear still performs the strongest.

\emph{To analyze}, it is shown in Table \ref{tab:overlapp} how the four languages are intersected with each other.
We introduce the concept of the language-degree for a phone in multilingual training.
Language-degree 4 means that the phone is shared by all the 4 languages, there are 18 such phones. Language-degree 1 means that the phone is language-unique, belonging to only one language. 
Italian has the smallest number of language-unique phones among the four languages. Many phones in Italian are also shared by other languages. Also note that Italian has the smallest amount of training data, as can be seen from Table \ref{tab:datainfo}.
These may explain the most significant benefit for Italian from multilingual training. For Italian, finetuned multilingual JoinAP-Nonlinear reduces the WER by 17\% again the monolingual Flat-Phone baseline. Also the gain by JoinAP-Nonlinear over JoinAP-Linear in Italian is also the largest (7\% without finetuning, 6\% after finetuning).
On the other hand, French has the largest number of language-unique phones among the four languages, and the training data size is larger.
This may explain the small improvement for French from multilingual training.

\begin{table}[]
\centering
\vspace{-1mm}
\caption{WER results (\%) for Polish in the crosslingual experiments. \#Finetune denotes the amount of data used in finetuning (0 means zero-shot).} 
\label{tab:polish}
\begin{tabular}{cccc}

\hline
\#Finetune  & Flat-Phone & JoinAP-Linear & JoinAP-Nonlinear\\
\hline
0          & 33.15 & 35.73 & 31.80 \\
10 minutes          & 8.70  & 7.50 & 8.10\\
\hline
\end{tabular}
\vspace{-3mm}
\end{table}

\begin{table}[]
\centering

\caption{WER results (\%) for Mandarin in the crosslingual experiments.}
\label{tab:mandarin}
\begin{tabular}{cccc}

\hline
\#Finetune  & Flat-Phone & JoinAP-Linear & JoinAP-Nonlinear\\
\hline
0          & 97.10  & 89.51 & 88.41 \\
1 hour          & 25.39  & 25.21 & 24.86\\
\hline
\end{tabular}
\vspace{-3mm}
\end{table}

\vspace{-2mm}
\subsubsection{Crosslingual experiment}

Crosslingual results for Polish and Mandarin are summarized in Table \ref{tab:polish} and Table \ref{tab:mandarin} respectively.
The two languages are representative in how much the testing language is overlapped with the training languages, or say in the other way, how many unseen phones are in the testing language, as seen in Table \ref{tab:unique}. Polish represents the much overlapping setting, while Mandarin the less overlapping setting. The results for the two settings are different, as detailed below.

For Flat-Phone, in order to calculate the phone probabilities for unseen phones from the new languages (Polish and Mandarin), the parameters connecting to unseen phones in the output layer of the DNN model are initialized randomly.
For JoinAP-Linear and JoinAP-Nonlinear, it is straightforward to calculate the phone embeddings for unseen phones, according to Eq. (\ref{eq:embedding}) and Eq. (\ref{eq:nonlinear}), once we obtain the phonological-vectors for those unseen phones.

\emph{In the zero-shot setting}, for both Polish and Mandarin, JoinAP-Nonlinear outperforms both JoinAP-Linear and Phone-Flat significantly and consistently.
The JoinAP-Linear performs worse than Flat-Phone in Polish. This is somewhat unexpected, which may reflect some instability of JoinAP-Linear.

\emph{In the few-shot setting}, for both Polish and Mandarin, JoinAP-Nonlinear still yields much better results than Phone-Flat; however, the superiority of JoinAP-Nonlinear over JoinAP-Linear seems to be weakened, as there are more training data from the target languages.
The JoinAP-Nonlinear performs better than JoinAP-Linear in Mandarin, while not in Polish.
10 minutes of transcribed speech in Polish may be rich enough for JoinAP-Linear to be well adapted, as indicated by the low WER.
As can be seen from Table \ref{tab:unique}, the number of unseen phones in Polish is far less than in Mandarin. This may also explains the good performance of JoinAP-Linear in Polish.
Remarkably, under a large amount of finetuning data for a testing language that is much overlapped with training languages, say, 1 hour for Polish, the performances of different models will tend to saturate and become less differed. So we use 10-min finetuning data for Polish for comparing different models. 10-min for Mandarin will yield results similar to zero-shot, so we use 1-hour for Mandarin few-shot.

\begin{table}[t]
\centering
\vspace{-4mm}
\caption{Statistics about Polish and Mandarin, including the number of IPA phone tokens in every language, and the number of unseen phones.
}
\label{tab:unique}
\begin{tabular}{ccc}
\hline
Language  & \#Phones & \#Unseen phones\\
\hline
Polish         & 46  & 18  \\
Mandarin          & 96  & 79 \\
\hline
\end{tabular}
\vspace{-4mm}
\end{table}

\begin{table}[t]
\centering
\vspace{-5mm}
\caption{Detailed explanation of Fig. \ref{fig:tsne}. We list the IPA phones in the colored circles from different methods.}
\label{tab:tsne}
\begin{tabular}{c|ccc}

\hline
Method  & Color & Feature & Phones\\
\hline
\multirow{3}{*}{Flat}   &\multirow{2}{*}{Green} & Retroflex & \textrtailz\  \textrtaild\textrtailz\ t\textrtails\\
  & & Palatalized  &  \textscriptg\textsuperscript{j}\ k\textsuperscript{j}\ \textctc\textsuperscript{j}\ v\textsuperscript{j}\ b\textsuperscript{j}\ x\textsuperscript{j}\ p\textsuperscript{j}\  \\
                                        \cline{2-4}
                                        & Red & Fricative  & b\ t\ p\ d\ k\ \\
\hline
\multirow{8}{*}{Linear}   &\multirow{4}{*}{Green} & Alveolo-palatal & \textctz\ \textctc\ \textdctzlig\ \texttctclig\\
  & & Velar  & \textipa{N}\ \textscriptg\ k\ x\ \textscriptg\textsuperscript{j}\ k\textsuperscript{j}\ x\textsuperscript{j}  \\
  & & Alveolar  & \textdzlig\ \texttslig\ n\ d\ t\ r\ s\ z\ l  \\
  & & Retroflex  & \textrtails\ \textrtailz\  \textrtaild\textrtailz\ t\textrtails \\
                                          \cline{2-4}
                                          &\multirow{2}{*}{Red} & Plosive  & x\ v\ x\textsuperscript{j}\ v\textsuperscript{j}\ z\ s\ f\ \textrtailz \ \textrtails \\        
& & Fricative & \textscriptg\ k\ p\ \textscriptg\textsuperscript{j}\ k\textsuperscript{j}\ p\textsuperscript{j}\ b\ b\textsuperscript{j}\ t\ d  \\
                                          \cline{2-4}
                                          &\multirow{2}{*}{Yellow}
& Close  &  i\ u\ w\ \textbari\ \~\textbari\\
 & & Open/Open-mid & a\ \textepsilon\ \textopeno\ \~\textopeno  \\
                                          \hline
\multirow{8}{*}{Nonlinear}&\multirow{3}{*}{Green} & Alveolo-palatal & \textctz\ \textctc\ \textdctzlig\ \texttctclig  \\
& & Velar  & \textipa{N}\ \textscriptg\ k\ x\ \textscriptg\textsuperscript{j}\ k\textsuperscript{j}\ x\textsuperscript{j}  \\
& & Alveolar  & n\ d\ t\ r\ s\ z\ l  \\       
                                          
                                          \cline{2-4}
                                          &\multirow{3}{*}{Red} & Affricate  & \textdctzlig\ \texttctclig\ \textdzlig\ \texttslig  \\        
& & Plosive  & x\ v\ x\textsuperscript{j}\ z\ s\ f  \\  
& & Fricative & p\ p\textsuperscript{j}\ b\ b\textsuperscript{j}\ t\ d  \\
                                          \cline{2-4}
                                          &\multirow{2}{*}{Yellow} 
& Close  &  i\ j\ \textbari\ \~\textbari\\
 & & Open/Open-mid & a\ \textepsilon\ \~\textepsilon\ \textopeno\ \~\textopeno  \\
\hline
\end{tabular}
\vspace{-5mm}
\end{table}

To further understand our phonological-vector based phone embeddings, we apply t-SNE \cite{tsne2008} to draw the 2048-dimensional phone embeddings on a 2-dimensional map. Fig. \ref{fig:tsne} shows the maps of the 46 phones in Polish, obtained from the un-finetuned multilingual models. 
It seems that we can hardly find many sensible groupings for flat phone embeddings from Fig. \ref{fig:tsne}(a).
But for JoinAP-Linear and JoinAP-Nonlinear phone embeddings, the maps reflect more notable groupings, where similar phones are found to gather together in the maps. We use red, green and yellow circles to indicate the phones with the same manner of articulation, place of articulation and vowel height respectively. Their detailed IPA features are listed in Table \ref{tab:tsne}. The figures clearly show that the JoinAP based phone embeddings indeed carry phonological information, which could help zero-shot learning. 
Moreover, it can be seen that the vowels are located in the top-right corner in Fig. \ref{fig:tsne}(c), while the vowels are separated in two corners in Fig. \ref{fig:tsne}(b). And the within-class scattering of (b) seems to be larger than (c). These observations could reflect the superority of JoinAP-nonlinear over JoinAP-Linear.

\section{Conclusion}
In this work, we propose the JoinAP method to join phonology driven phone embedding (top-down) and DNN based acoustic feature extraction (bottom-up).
We apply linear or nonlinear transformation of phonological-vectors to obtain phone embeddings, and compare to the traditional method using flat phone embeddings.
In the multilingual and crosslingual experiments, JoinAP-Nonlinear generally performs better than JoinAP-Linear and the traditional flat-phone method on average. 
The improvements are generally the most significant for those target languages such as Italian in our multilingual experiments and Polish in our zero-shot crosslingual experiments, due to their data-scarce and high language-degrees of their phones (i.e., being well shared by other languages), and become weak for those target languages when they become data-rich such as French in our multilingual experiments and Polish in our few-shot crosslingual experiments.
In summary, the JoinAP method provides a principled approach to multilingual and crosslingual speech recognition.
Some promising directions include exploring DNN based phonological transformation, and pretraining over increasing number of languages.



\bibliographystyle{IEEEbib}
\bibliography{strings,refs}

\end{document}